# Topological Navigation of Simulated Robots using Occupancy Grid

**Richard Szabo**

Eotvos Lorand University , Budapest, Hungary
Department of General Computer Science,
Department of History and Philosophy of Science,
rics@inf.elte.hu

*Abstract: Formerly I presented a metric navigation method in the Webots mobile robot simulator. The navigating Khepera-like robot builds an occupancy grid of the environment and explores the square-shaped room around with a value iteration algorithm. Now I created a topological navigation procedure based on the occupancy grid process. The extension by a skeletonization algorithm results a graph of important places and the connecting routes among them. I also show the significant time profit gained during the process.*

## 1. Introduction

Mobile robotics and robot navigation is a growing area of scientific research. Without navigation the creation of self-propelled, household machines, guard robots, or planet surveyors is beyond imagination.

Robot simulators are useful designing and analizing tools of the navigation research area since learning can be more effective in the computer than in the real world.

In this paper I present a method for building topological navigation graph on the top of an occupancy grid. First of all I show in brief the creation of an occupancy grid and the exploration with value iteration. Then I focus on the necessary steps of the composition of the graph and the environment exploration utilizing the evolved graph.

The primary tool for the experiments is the Webots mobile robot simulator. This tool is capable of imitating almost any type of mobile robots including wheeled, legged, and flying models (Michel, O., 2004).

This project is part of my Ph.D. research with the main aim of the investigation of mobile robot navigation. After I was the runner up of the 1st Artificial Life Creators Contest organized by Cyberbotics Ltd. in 1999, I won the second contest in 2000 and obtained the simulator license as the first prize. Details of the competitions are discussed in (Szabó, R., 2001).

## 2. Previous work

My former goal was to create a metric navigation module for a modified Khepera robot in the Webots simulation environment, that is to say I focus on metric spatial properties of objects like distances, and coordinates. The developed robot has to build a cognitive map — "a view from above" — of small rooms while it visits every reachable location (Szabó, R. 2003). Fig. 1., Fig. 2. show some typical experimental area.

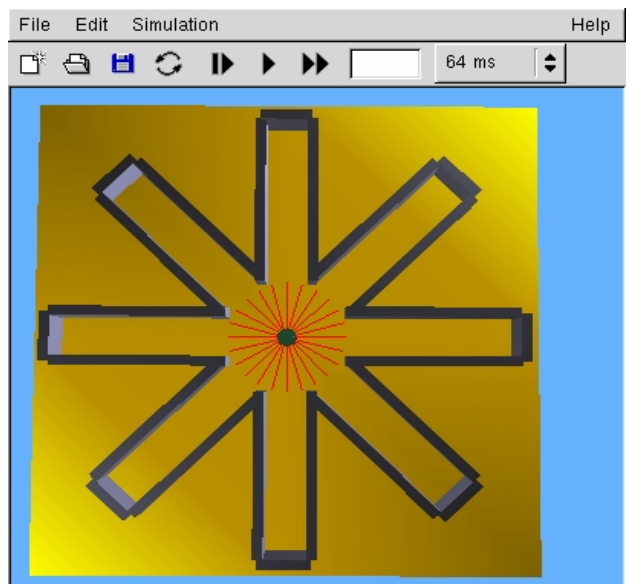

Fig. 1. Radial maze

In this experiment I modified the selected Khepera robot. The 8 infra-red distance sensors were changed to 24

sonars with a sensing range of 15 cm to facilitate the perception of the environment.

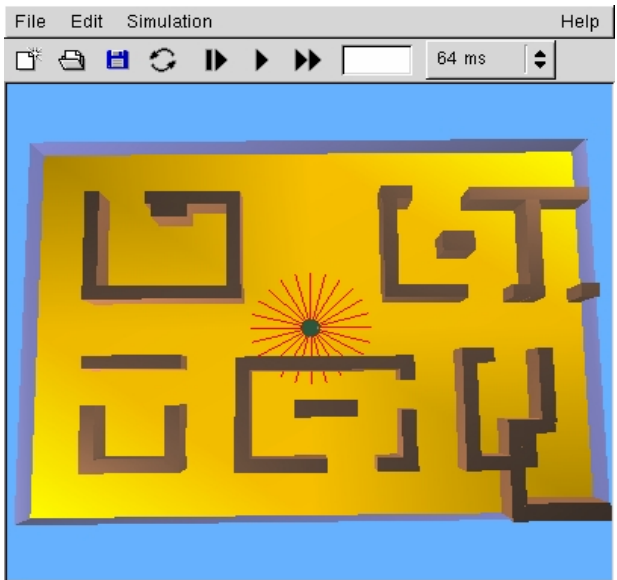

Fig. 2. AAAI contest maze

The adopted method of metric navigation is based on the occupancy grid model pioneered by Moravec and Elfes (Moravec, H. P. & Elfes, A., 1985 and Elfes, A., 1989). This general structure in two dimension manages a tesselation of the plane in cells. Each cell of the occupancy grid contains a probability value which is an estimation that the represented position is occupied by some object. After the investigation of other probabilistic navigation possibilities like Kalman-filter or expectation maximization (Thrun, S., 2002), I have chosen this grid technique because it is relatively simple to implement and because of its iterative nature.

The important steps of the map building, in accordance with Thrun's work (Thrun, S., 1998), are the following:

- sensor interpretation
- integration over time
- pose estimation
- global grid building
- exploration

*2.1. Occupancy grid creation*

The sensor interpretation is the first phase in the creation of the occupancy grid based navigation. During this process the 24 sonar scalar values are converted to local occupancy values around the robot. The conditional probabilities of grid cells are determined by a predefined conversion function: probabilities are high *at* the point of a measurement, and are low closer to the robot.

Since different sonar measurements give different values for a grid cell because of noise and changing viewpoint, it is important to integrate the conditional probabilites of distinct moments. Using the assumption of the indepence of measurements — that generally does not hold — and the Bayes theorem, incremental calculation of occupancy grid values is possible, sonar scans can be "concatenated" to previous experiences.

After the local grid is created around the robot its values have to be merged into the global grid. Beyond the coordinate transformation between the grids we need a global position where the local grid can be integrated. Robot position estimation is not an immanent property of the occupancy grid technique so an accepted method is to use a position estimation method like odometry — continuous calculation of changes of the robot pose — combined with correction of errors accumulated by sensor and motor noise (Borenstein, J., 1995). Since my main focus is the occupancy grid the simulator provides the position to step across this problem.

Fig. 3. shows the occupancy grid of a maze during the process of the exploration.

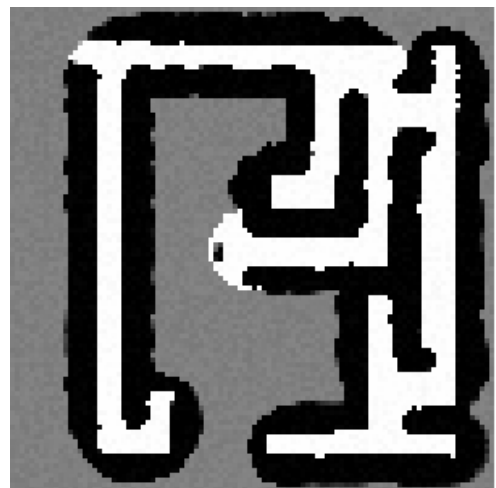

Fig. 3. Occupancy grid of a maze

*2.2. Iterative evaluation*

After the robot is ready to create the environment map, a driving force is needed to urge the robot to explore all the reachable places, otherwise it would wander randomly. For this reason we implemented a variant of value iteration. This method is well-known in the domain of reinforcement learning (Sutton, R. & Barto, A., 1998).

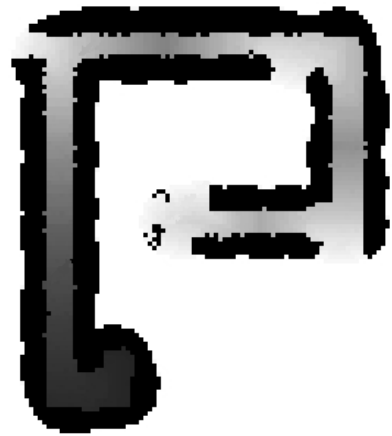

Fig. 4. Cost matrix of the maze

The selected algorithm helps to find the minimum cost-path to unexplored regions of the occupancy grid. A cost matrix is calculated iteratively and after convergence for every occupancy grid cell the cost of travelling to an unexplored grid cell from the actual cell is given (Fig. 4). Exploration direction is then a resultant of the cost matrix, the actual direction of the robot, and an obstacle-avoidance behaviour.

**3. Building a topological graph from occupancy grid**

Exploration using value iteration is a very time-consuming task. Values of the cells of the cost matrix are calculated by a process which scans through the whole matrix many times. Furthermore the next exploration direction is based on this gradient map and it does not take into account the constraint of the robot dynamism, sometimes resulting a fairly clumsy movement.

Accordingly it seems a natural improvement to replace the value iteration module with a topological graph. This graph emphasizes the links between landmarks, the possibility to move from one place to another. Edges represent traversable corridors of the environment and nodes are the crossings or end points. Navigation using the graph is much faster since its size is some order of magnitude smaller than of the cost matrix. Chapter 2 of my book (Szabó, R. 2001) compares metric and topological navigation in detail.

There are many different ways of creating a navigation graph using a metric map. Skeletonization, Voronoi-diagrams, matching opposite contours, sparse pixel approaches are among the possibilities (Thrun, S., 1998 and Tombre, K. & Ah-Soon, C. & Dosch, P. & Masini, G. & Tabbone, S., 2000). In any case, the occupancy grid can be viewed as a two-dimensional greyscale image of the environment, hence digital image processing methods are valid approaches (Elfes, A. 1989).

Since I selected skeletonization, steps of topological navigation using the occupancy grid are the following:

- skeletonization
- chaining the skeleton to form edges
- graph optimization
- navigation using the graph

*3.1. Skeletonization*

I decided to produce the skeleton of the explored and unoccupied region of the environment. At the end of the process skeleton points are those places where the robot is hopefully not blocked by any obstacles.

For this reason I utilized medial axis transform (MAT) (Borgefors, G., 1986). An interior point of the shape belongs to the medial axis if this point lies at the same distance from two or more nearest contour points. Unfortunately one drawback of MAT appeared during my tests: medial axis of discrete objects and shapes — like the discrete occupancy grid to be projected — may be disconnected. This deficiency is not acceptable in our case since the resulting skeleton has to contain all connected routes among landmarks of the environment.

As a second attempt instead of using medial axis transform I applied a thinning algorithm to "peel the union", in other words I iteratively shrinked the object to its one pixel wide skeleton (Jain, A. K., 1989). During this process the border pixels are deleted successively while topology and morphology of the object is preserved, that is to say no pixels are deleted at the end of a line or at the connection of two regions.

The thinning algorithm works as it is described in Algorithm 1. Labeling of pixels around the actual pixel ($P_1$) advances counter-clockwise.

---

$Z0(P_1)$ - the number of zero to nonzero translations in the sequence $P_2, P_3, P_4, P_5, P_6, P_7, P_8, P_9, P_2$
$NZ(P_1)$ - the number of nonzero neighbours of $P_1$
Steps:
1. Scan through all the points of the image.
2. Calculate $Z0(P_1)$, $NZ(P_1)$, $Z0(P_2)$, $Z0(P_4)$, for all points.
3. Delete $P_1$ if the conditions simultaneously satisfied:
$$2 <= NZ(P_1) <= 6,$$
$$Z0(P_1) = 1,$$
$$P_2 * P_4 * P_8 = 0 \text{ or } Z0(P_2) <> 1$$
$$P_2 * P_4 * P_6 = 0 \text{ or } Z0(P_4) <> 1$$

---

Algorithm 1. The thinning algorithm

Fig. 5. is an example of the result of the skeletonization process using the thinning algorithm.

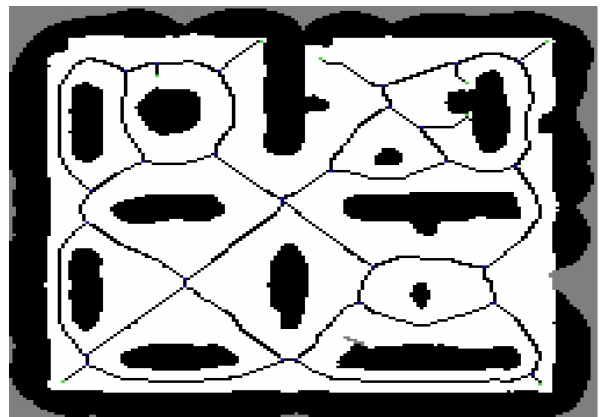

Fig. 5. Skeleton of an office

*3.2. Chaining*

Navigation on the skeleton of the explored and unoccupied territory is possible and can be more effective than the calculation of the cost matrix of the value iteration because thinning results a data compression. Nevertheless it is advisable to use the skeleton as a basis for further processing.

Skeleton of the explored region is a set of pixels, this structure can be transformed to a graph. First of all, those points have to be determined where skeleton branches meet. These pixels are the nodes, otherwise they are the crossing points of corridors. After I have selected the nodes I cycle through the skeleton branches. This procedure issues in chains, what are pixel sequences from node to node or from node to skeleton end point (Tombre, K. & Ah-Soon, C. & Dosch, P. & Masini, G. & Tabbone, S., 2000). Algorithm 2. reveals the main structure of the procedure.

```
while there are nodes left do
  c = newChain()
  while there are non-null neighours left do
    if not found getNonNode4Neighbour(q) then
      if not found getNode4Neighbour(q) then
        if not found getNonNode8Neighbour(q) then
          if found getNode8Neighbour(q) then
            append(q,c)
          end
          endChain(c)
        else
          append(q,c)
        end
      else
        append(q,c)
        endChain(c)
      end
    else
      append(q,c)
    end
  end
end
```

Algorithm 2. Excerpt of the chaining algorithm

The first draft of the graph is calculated during the chaining process. Skeleton nodes and end points take part in the graph as nodes. Graph edges connect those nodes between which a chain exists.

During my investigation it turned out that the cited algorithm has two minor problems that, in special cases, corrupts the graph. On Fig. 6. and Fig. 7. chain creation starts from nodes (marked by 'o') and cycles through all the neighbours of the node (marked by 'x'). Non-node elements are cancelled after they take part in a chain.

First problem rises in situations similar to the one shown on Fig. 6. Pixel x marked by 1 (x-1) is cancelled during the chain creation starting from x-2. In the next step — since all the neighbours of nodes have to be processed — chain creation tries to start from an already cancelled node: x-1.

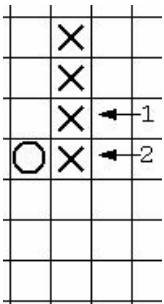  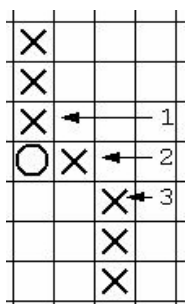

Fig. 6. Chain problem 1    Fig. 7. Chain problem 2

Another problem is indicated on Fig. 7. If the chain creation starts from x-2 then in the next step the search should turn to x-3 and chain the pixels downwards. However there is no explicit constraint in the algorithm to prevent the continuation after x-2 in the direction of x-1, what is obviously wrong, since it leaves x-3 without a connection to the node. After I corrected these mistakes the chaining algorithm created the draft of the navigation graph.

*3.3. Graph optimization*

First version of the graph is not applicable to navigate because chains may ramble far away from edges and if the robot simply follows the way of an edge it could meet with obstacles.

To cope with this problem it is possible to recursively split the edge in question and ensure that the new particles track the slues of the chain better. There are two different algorithm-family for this approximation.

Wall and Danielsson calculate the algebraic surface between the edge and the chain (Wall, K. & Danielsson, P.-E., 1984). The iterative computation is performed by determining the sum of successive triangles. If the size of the surface exceeds a certain threshold then splitting of the edge is necessary.

Rosin and West's algorithm measures the maximal distance between the edge and the chain (Rosin, P. L. & West, G. A., 1989). This method splits the edge at its maximum deviation point recursively until all the created new edges are acceptable approximations of the chain (Fig. 8.).

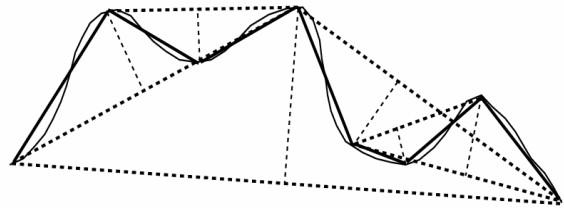

Fig. 8. Splitting (taken from the slides of Tombre, K. & Ah-Soon, C. & Dosch, P. & Masini, G. & Tabbone, S., 2000)

As a comparison of the methods Tombre, K. & Ah-Soon, C. & Dosch, P. & Masini, G. & Tabbone, S., 2000 states that Wall and Danielsson can be implemented very efficiently but, on the other hand, it is less accurate than Rosin and West's method. Additionally, the second mentioned algorithm may split up edges into small pieces near junctions.

Since I would like to use the topological graph for navigation at the end, it is important that edges do not cross or reach obstacles and walls. In other words, fidelity of the graph to the calculated chain is important so I have chosen and implemented Rosin and West's algorithm. The procedure is described in Algorithm 3.

When the recursive splitting is finished pruning of edges is useful especially near to unexplored regions. Otherwise, if the robot simply moves to an end node where unexplored territory is nearby, then accidentaly it could run into a wall.

```
split_edge(graph,start_point,end_point) {
  while chain is not finished do
    act_point = get_act_point(chain)
    h = height(start_point,end_point,act_point)
    if h > LIMIT then
      delete_edge_from_graph(graph,start_point,end_point)
      add_node_to_graph(graph,act_point)
      add_edge_to_graph(graph,start_point,act_point)
      add_edge_to_graph(graph,act_point,end_point)
      split_edge(graph,start_point,act_point)
      split_edge(graph,act_point,end_point)
    end
  end
}
```
Algorithm 3. Algorithm of Rosin and West

Fig. 10. shows the optimized graph of Fig. 9. after recursive edge splitting and pruning.

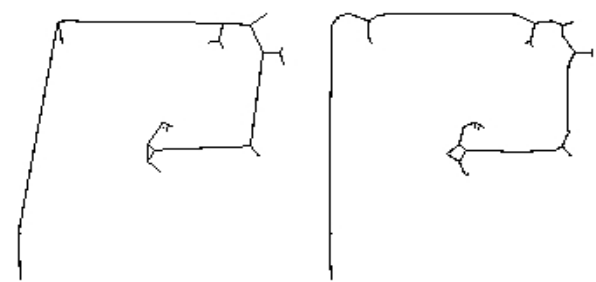

Fig. 9. Graph after chaining   Fig. 10. Optimized graph

*3.4. Navigation*
When creation of the graph of the explored and free region is complete, the robot has to determine the next exploration direction. Generally the robot is aimed to sweep through all the reachable places of the terrain. This is why those nodes of the graph can be considered as goal nodes where unexplored region is close.
To localize these elements I performed a general $A^*$ algorithm (Futó, I., 1999). This classical algorithm finds the shortest path from the predefined start node of the graph to a goal node. Start node of the graph in our case is the actual position of the robot. The $A^*$ algorithm then calculates the shortest path from the actual position to a node where exploration could be fruitful.
Using the shortest path as a list to be processed, the robot can turn to the next node of the graph in the list and move directly ahead while it does not reach the last node in the list.
Besides the topological graph and the $A^*$ algorithm the final robot movement is comprised another behaviour pattern as well. The role of this normal move module is to stimulate the robot straight ahead "en plaine air", and it also ensures obstacle avoidance motion in case of necessity. Since the generation of the topological graph is time-consuming, this job is not done continuously. When the normal move module does not explore efficiently, in other words the explored surface does not grow enough, creation of the graph takes place and navigation is governed by the $A^*$ algorithm. This alternating comportment incorporates the advantages of the two behaviour modules.

## 4. Results

During our research I created a topological navigation method based on occupancy grid in the simulation environment of Webots. Using topological graph instead of value iteration for the determination of exploration direction seems to be a beneficial modification. On one hand it approximates better the nature of the navigation. On the other hand the new algorithm performs better.
The two evolved robot controllers were tested in five different environments in several experiments from various starting points. The environments were selected to cover a wide range of possible situation that could arise during map-building.
The terrains were the following: an open area with some round obstacles, a radial maze taken from Csányi, V., 1994, well-known in cognitive map researches (Fig. 1.), a maze, an office-like room which was one of the fields of the Artificial Life Creators Contest, and a labyrinth used at the 1994 AAAI autonomous mobile robot competition (Fig. 2., Thrun, S., 1998).
The open area is 1 $m^2$, the AAAI maze is 1.85 $m^2$, while the others are 2.25 $m^2$. Five attempts were performed in every field with both algorithm. The robot could explore all the environments by the two methods.
In the small and easily solvable open area the robot spends 8 and 6.4 minutes on an average in robot performance time using value iteration and topological graph respectively. Radial maze does not cause any difficulties for the two programs, both solve it in around 6 minutes on an average.
The most significant advance can be reached in the office environment: the 20 minutes time drops to 12.4 minutes. In the maze the time profit is smaller: the 22 minutes of value iteration is reduced to 14.5 minutes. The AAAI contest environment is easier to solve than the maze, hence time frames of value iteration and graph navigation are 14 and 11.7 respectively.
These results are collected in Table 1.
The acceleration between the two methods is a consequence of the smaller number of entities with which the algorithms have to deal (Table 2.).

|  | Value iteration (min) | Topological graph (min) |
|---|---|---|
| Open room | 8 | 6.4 |
| Radial | 6.3 | 6 |
| Office | 20 | 12.4 |
| Maze | 22 | 14.5 |
| AAAI contest | 14 | 11.7 |

Table 1. Time comparison of the navigation methods

There are between 11600 and 28900 pixels in the cost matrix of the value iteration, and the number of graph nodes are between 20 and 120, depending on the size and the complexity of the environment.

|  | Value iteration (pixels) | Topological graph (nodes) |
|---|---|---|
| Open room | 12800 | 50 |
| Radial | 11600 | 20 |
| Office | 28900 | 110 |
| Maze | 28900 | 120 |
| AAAI contest | 23700 | 105 |

Table 2. Number of entities in the navigation methods

## 5. Conclusions

This paper presents a method to build a topological graph for navigation based on occupancy grid. Besides the fact that already known algorithms are used, significantly better accomplishments related to the pure occupancy grid method justify this navigation approach.
On one hand the number of manipulated entities —pixels for the value iteration, and graph nodes for the topological navigation — differ in the two approaches. This gap is more than two orders of magnitude, so the graph navigation dramatically reduces the need for resources, especially the need for memory.
On the other hand better total exploration time can be achieved with the newer control procedure. Differences in the acceleration among various test fields follow from the fact that the graph mostly helps in elongated parts of the territory and at the connections of the large spaces. Open spaces are easily explorable by random obstacle avoidance so the necessary time for open room and radial maze is not diminished essentially. For the maze, the office, and the AAAI contest environment the effects are easily recognizable, since time profit exceeds 20%.

## 6. Future work

There are quite many different ways of continuing the research. Some of them are mentioned below:
- Testing the algorithms in real robot.
- Higher level task can be performed by the robot after successful exploration.
- Moving around in dynamic environments is a serious challenge, this extension would make the problem more interesting.
- Using position estimation may make the robot fully automate.
- Introduction of new sensor types especially video cameras may enhance the occupancy grid creation and position estimation as well.

## 8. Acknowledgement
The author wishes to thank György Kampis for his useful suggestions, and András Salamon for his valuable remarks.